\RequirePackage{fix-cm}
\documentclass{svjour3}                     
\smartqed  
\usepackage{graphicx}
\usepackage{multirow}
\usepackage[ruled,vlined]{algorithm2e}
\usepackage{amsmath}
\usepackage[square,numbers]{natbib}
\begin{document}
\title{Semantic Segmentation of Surface from Lidar Point Cloud
}


\author{Aritra Mukherjee$^1$         \and
        Sourya Dipta Das$^2$         \and
        Jasorsi Ghosh$^2$            \and
        Ananda S. Chowdhury$^2$      \and
        Sanjoy Kumar Saha$^1$
}
\authorrunning{A Mukherjee et al.} 

\institute{$1$ Department of Computer Science \& Engineering, 
Jadavpur University\\
Kolkata, India\\
$2$ Department of Electronics \& Telecommunication Engineering, 
Jadavpur University\\
Kolkata, India
}

\date{Received: date / Accepted: date}

\maketitle

\begin{abstract}

In the field of SLAM (Simultaneous Localization And Mapping) for robot navigation, mapping the environment is an important task. In this regard the Lidar sensor can produce near accurate 3D map of the environment in the format of point cloud, in real time. Though the data is adequate for extracting information related to SLAM, processing millions of points in the point cloud is computationally quite expensive. The methodology presented proposes a fast algorithm that can be used to extract semantically labelled surface segments from the cloud, in real time, for direct navigational use or higher level contextual scene reconstruction. First, a single scan from a spinning Lidar is used to generate a mesh of subsampled cloud points online. The generated mesh is further used for surface normal computation of those points on the basis of which surface segments are estimated. A novel descriptor to represent the surface segments is proposed and utilized to determine the surface class of the segments (semantic label) with the help of classifier. These semantic surface segments can be further utilized for geometric reconstruction of objects in the scene, or can be used for optimized trajectory planning by a robot. The proposed methodology is compared with number of point cloud segmentation methods and state of the art semantic segmentation methods to emphasize its efficacy in terms of speed and accuracy. 


\end{abstract}

\section{Introduction}
\label{intro}

3D mapping of the environment is an important problem for various robotic applications and is one of the two pillars of SLAM (Simultaneous Localization And Mapping) for mobile robots. Various kinds of sensors are in use to achieve the goal. Stereo vision cameras are one of the cheapest solution and works satisfactorily for well lit, textured environments but fails for the places lacking unique image features. Structured light and Time of Flight (ToF) cameras gives real time depth information for pixels in the image of a scene (RGBD) and is good for indoor usage. But in the presence of strong light {\it i.e.} in outdoor environments, its efficiency suffers a lot. Lidar is the primary choice for mobile robots working in the environments with diverse illumination and structural features. Lidar works on the principle of measuring time of flight of short signature bursts of laser that can be filtered out from other forms of radiations. As a result its robustness and range are increased. The downside of Lidar is its low resolution and thus a fair amount of computation is needed to extract usable information from Lidar data. This computational load is one of the deterrents for its usage. Thus there is scope for research in formulating efficient algorithms for Lidar point cloud processing. 

The present work exploits the working principle of spinning Lidar to generate a near accurate mesh of the environment in an online fashion. The mesh is built on a subsampled cloud to increase speed of operation. The mesh is used to estimate surface normal of the subsampled points. On the basis of angle between surface normals a simple threshold based graph traversal approach is used to generate surface proposals. A binned histogram of surface normals is used as feature to train and use a Random Decision Forest (RDF) classifier to estimate a semantic surface label for the segment. Such semantic segments can be further utilized to estimate geometric models of object parts in the scene for scene reconstruction or can be directly used for smarter navigation of mobile robots. 

The present paper is divided into following sections. Section~\ref{sec:prev_works} discusses some major works regarding surface segmentation of 3D point clouds. Section~\ref{sec:method} elaborates about the proposed methodology. Section~\ref{sec:results_comp} sheds light on the results of the proposed methodology along with its comparison with other relevant works. Finally, Section~\ref{sec:conc} concludes the work.

\section{Previous Works}
\label{sec:prev_works}
The field of Lidar point cloud segmentation is comparatively new. Though segmentation of dense point cloud obtained from meticulous scanning of 3D models is an old problem, fast segmentation of sparse point cloud for robotic applications gained impetus in recent years. Some of the important works regarding the problem are as follows.

According to a survey~\cite{Nguyen2013}, the classical approaches for point cloud segmentation can be grouped as: edge based methodologies~\cite{Bhanu1986}, region based methodologies~\cite{Jiang1996}~\cite{Vo2015}~\cite{Li2017}~\cite{Bassier2017}, derived attributes based methodologies~\cite{Zhan2009}~\cite{Ioannou2012}~\cite{Feng2014}~\cite{Hackel2016}, model based methodologies~\cite{Tarsha-Kurdi2007} and graph based methodologies~\cite{Rusu2009}~\cite{Golovinskiy2009}~\cite{Landrieu2018}~\cite{Ben-Shabat2018}.

Vo et al.~\cite{Vo2015} proposed an octree based region growing method with segment refinement. Bassier et al.~\cite{Bassier2017} further improved it with the help of Conditional Random Field. Variants of region growing approach was proposed earlier by Jiang et al.~\cite{Jiang1996} and later Li et al.~\cite{Li2017} have followed similar approach to work on the range image generated by 3D point clouds. For segmentation of unorganized point clouds, a Difference of Normal (DoN) based multiscale saliency feature was considered by Ioannou et al.~\cite{Ioannou2012}. For Lidar, it is often easier to process the point cloud if it is represented in polar or cylindrical coordinates rather than cartesian coordinates. Line fitting in segments to a point cloud represented in cylindrical coordinates was proposed in the work of Himmelsbach et al.~\cite{Himmelsbach2010}, which was further filtered to extract ground surface. Using undirected graph as mesh building and subsequent estimation of ground surface by local normal comparison was also proposed by Moosman et al.~\cite{Moosman2009}. A fast instance level LIDAR Point cloud segmentation algorithm was proposed by Zermas et al.~\cite{Zermas2017}. It deals with deterministic iterative multiple plane fitting technique for fast extraction of the ground points and it is followed by a point cloud clustering methodology named Scan Line Run (SLR). 

Point cloud segmentation methods using deep learning is a recent trend and there are few works in this direction. PointNet~\cite{Qi2017} uses the 3D sliding window approach for semantic labelling of points. It assumes that local features in the neighbourhood window of a point is enough to estimate its semantic class. PointNet++~\cite{Qi2017pnpp} further refined the method by applying pointnet on a nested partitioning of the cloud in a hierarchical fashion. PointCNN~\cite{li2018pointcnn} first learns an $\mathcal{X}$ transformation that weighs input features describing a point and permutation of points that makes it ordered. Then it applies Convolutional Neural Network (CNN) on the ordered points for semantic labelling. PointSIFT~\cite{jiang2018pointsift} is a preprocessor for various deep net based approaches that applies a 3D version of SIFT(Scale Independent Feature Transform) on pointcloud to make it orientation and scale invariant. Thereby it enables the training of network with fewer instances. Point Voxel CNN~\cite{liu2019point} uses voxel based approach for convolution. Thus, it saves time and memory that is wasted on structure mapping of point clouds. DeepPoint3D~\cite{srivastava2019deeppoint3d} uses multi-margin contrastive loss for discriminative learning so that directly usable permutation invariant local descriptors can be learnt. Most of the approaches uses convolutional neural network but graph neural network (GCN) can also be used for semantic segmentation of 3D point cloud~\cite{li2019deepgcns}. In GCN the convolution happens on subgraphs rather that local patches and is thus useful for non-euclidean data such as point cloud. Very recently PolarNet~\cite{zhang2020polarnet} used a polar grid based representation for online semantic segmentation of point cloud from spinning Lidars. 

In the earlier works, surface normal is estimated from point neighbourhood which is determined by tree based search. For sparse point cloud, like that from Lidar, this approach is time consuming and prone to errors. This motivated us to develop a fast online meshing of point cloud which can be used for surface normal estimation of points during the scan. An earlier attempt~\cite{mukherjee2019fast} used this normal as the feature for surface segment propagation resulting in unlabelled segmentation. For semantic labelling recent works have heavily relied on deep learning. On the other hand we have gone with traditional machine learning methods for estimating surfaces that can be used for surface fitting for scene reconstruction using vector models. The semantic surfaces, especially the ground plane detected out of the present form is useful in robot navigation purposes. 

\section{Proposed Methodology}
\label{sec:method}

\begin{figure}[ht]
\centering
\includegraphics[width=\textwidth]{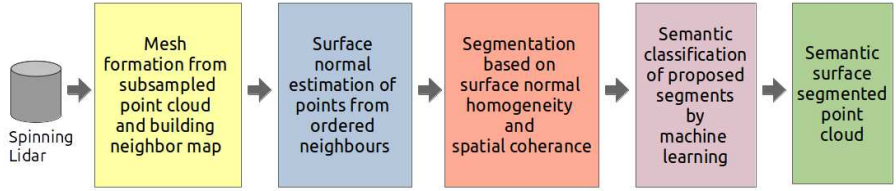}
\caption{Process flow for the entire system, the first two stages can be made online if data is sampled from Lidar on the fly.}
\label{fig:lidar_block_dia}
\end{figure}

The process to semantically segment the surfaces from Lidar data consists of four major stages as shown in Fig.~\ref{fig:lidar_block_dia}. Part of the system was designed in our previous work~\cite{mukherjee2019fast}. Previously only surface segments were generated from point clouds. In the present form, semantically labelled surface segments are generated. Other than the first stage which remains unchanged, all stages are updated. A new stage of semantic segmentation is also added. 

The first stage forms the mesh from subsampled point clouds. The subsampling is based on skipping regular number of vertical scans while the Lidar makes a single spin. The second stage estimates a surface normal for the subsampled points using the local mesh information. It should be noted that in actual use these two stages can be performed in an online fashion. The third stage is tasked with formation of segment proposals based on local distribution of surface normals. The number of proposals generated depends on the nature of the point cloud and also a control parameter discussed later. The final stage processes the proposed segments with a machine learning based classifier to assign a semantic label to that. The entire process is independent of Lidar orientation and scale, but only applies for spinning Lidars.

\begin{figure}[ht]
\centering
\includegraphics[width=\textwidth]{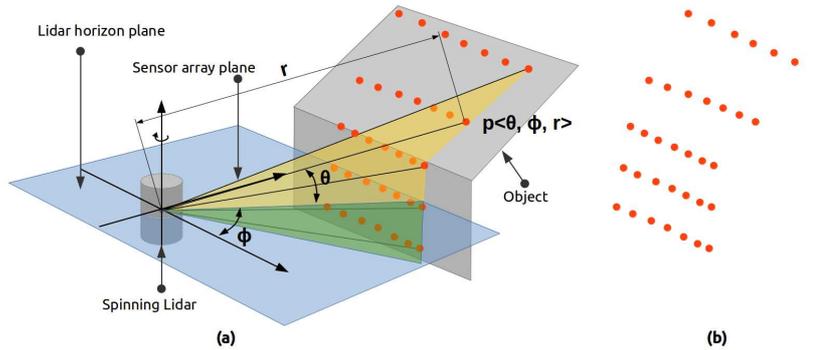}
\caption{(a) A schematic diagram~\cite{mukherjee2019fast} showing the formation of point cloud by Lidar and (b) the resultant point cloud.}
\label{fig:lidar_op}
\end{figure}

A spinning Lidar works by spinning a vertical array of units that measure distance by laser sensor. The units individually measures distance by the time of flight method corresponding to a signature pulse of laser. Such pulse does not overlaps with secondary radiations present in the environment. Thus the data acquired is in a point cloud format in spherical coordinates. Each point in the cloud can be defined as $P = \{p(\theta , \phi , r)\}$ where, $\theta$ is the constant vertical angle of a sensor unit with respect to the plane perpendicular to the spinning axis, $\phi$ is the horizontal angle that varies when the array spins and $r$ is the distance measured by the laser sensor unit. If only a single spin is considered then this representation helps towards structuring of point cloud. For our methodology all $phi$ values are not considered and is thus subsampled at regular intervals. This can be done safely as the horizontal density of points in a spinning Lidar is quite high. Thus, by varying the sampling intervals, horizontal density of the point cloud can be varied. This step is also necessary as too close points can give a bad estimate for surface normal at a point due to sensor noise. Figure~\ref{fig:lidar_op}~\cite{mukherjee2019fast} elaborates the working principle of a spinning Lidar and the resultant point cloud formed for an object with multiple surfaces. The sampling interval is $2$ in this case {{\it }i.e.} every second point is sampled only for further computation. The points left out are labelled according to its nearest labelled point horizontally for generating dense surface segment proposals.

\subsection{Mesh Construction}
\label{subsec:mesh}
A crucial and significant part of the entire methodology is the novel fast mesh generation procedure. The fast operation of this stage ensures the speed and accuracy of the overall methodology. The mesh generation stage can be performed online as no global data is required. During the sweep the links between points are established in the following manner. Let a point be denoted as $p(\theta , \phi , r)$. Let the range of $\theta$ be $[\theta_0,\theta_n]$ which corresponds to $n+1$ vertical sensors in the array; and the range of $\phi$ be $[\phi_0,\phi_m]$ where $m+1$ is the number of times the sensor array is sampled uniformly during a single spin. The distance of a point is $r$ from the sensor unit and for computational purpose it is estimated that $r$ is the distance of the point from a virtual center from which all the sensor units are diverging. Let the topmost sensor in the array corresponds to angle $\theta_0$ and the sensors are counted in top to bottom order for an individual vertical sensor array. Also, let horizontal angle corresponding to the first shot in a spin be $\phi_0$. With these assumptions in place the mesh is constructed using the following rules.
\begin{enumerate}
    \item Form links between $p(\theta_i , \phi_j , r)$ and $p(\theta_{i+1} , \phi_j , r)$, $p(\theta_{i+1} , \phi_{j+1} , r)$, $p(\theta_i , \phi_{j+1} , r)$ for all points within range of $[\theta_0,\theta_{n-1}]$ and $[\phi_0,\phi_{m-1}]$. This covers most of the points during the spin.
    
    \item Form links between $p(\theta_n , \phi_j , r)$ and $p(\theta_n , \phi_{j+1} , r)$ where $j$ varies from $0$ to $m-1$. This takes care of the lowermost circle of the mesh as it was not handled by rule 1.
    
    \item At the end of the spin form links between $p(\theta_i , \phi_m , r)$ and $p(\theta_i , \phi_0 , r)$, $p(\theta_{i+1} , \phi_0 , r)$, $p(\theta_{i+1} , \phi_m , r)$. For the lowermost point form a link between $p(\theta_n , \phi_m , r)$ and $p(\theta_n , \phi_0 , r)$. This completes the cylindrical mesh.  
\end{enumerate}
It should be noted that rule 1 and 2 is started after two samples and is performed for all shots of vertical samples during the spin. Rule 3 is only applicable during the last shot of vertical sample. For a pair of points if $r$ for both the points are within the range of the Lidar then only linking is done for them. For points where all the points are present in a neighbourhood, six of the eight neighbours are linked to a point. If eight neighbours are linked, it will form crosses in the mesh thus violating its very definition of forming non-overlapping triangles. 

\begin{figure}[ht]
\centering
\includegraphics[width=\textwidth]{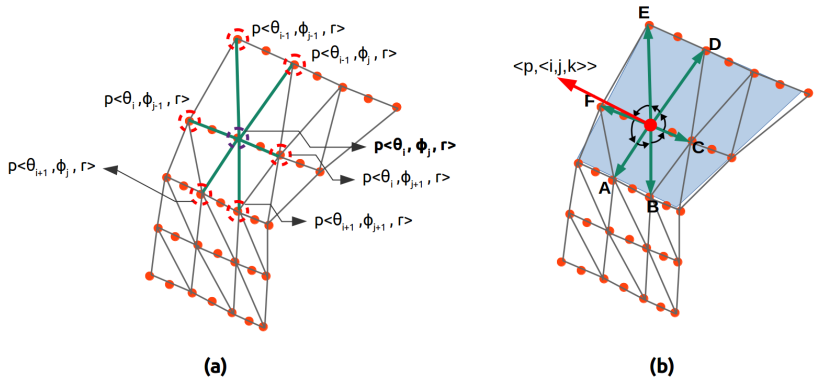}
\caption{(a) A schematic diagram~\cite{mukherjee2019fast} showing the formation of mesh on subsampled (factor 2) cloud with the neighbour definition of a point and (b) the normal formation from the neighbours.}
\label{fig:lidar_meshnorm}
\end{figure}

Figure~\ref{fig:lidar_meshnorm}(a)~\cite{mukherjee2019fast} shows the connectivity for a point which have six valid neighbours on the mesh. The mesh is stored in a map of vectors $M=\{ <p,v> \mid p \in P, v = \{ q_n \mid q_n \in P, n\leq6,\quad and\quad q_n\quad is\quad a\quad neighbour\quad of\quad p \} \}$ where each point is mapped to the vector $v$. $v$ contains the valid neighbors in an ordered fashion by traversing in an anticlockwise fashion from the bottom direction. This is performed either by an insertion sort or considering three consecutive vertical shots at once. For a point $p(\theta_i , \phi_j , r)$  with all six valid neighbours, the order of neighbours in $v$ is $p(\theta_{i+1} , \phi_j , r)$, $p(\theta_{i+1} , \phi_{j+1} , r)$, $p(\theta_i , \phi_{j+1} , r)$, $p(\theta_{i-1} , \phi_j , r)$, $p(\theta_{i-1} , \phi_{j-1} , r)$ and $p(\theta_i , \phi_{j-1} , r)$. Even if all valid neighbours are not present for a point, the order is maintained and is very crucial for normal estimation stage. The computational complexity of the mesh generation stage is $O(n_{sp})$ where $n_{sp}$ is the number of sub-sampled points. The mesh is generated during the spin of the Lidar and thus in actual scenario the computation time depends on angular frequency of the spinning Lidar and the regular interval or sub-sampling factor.

\subsection{Normal Estimation}
Surface normal is generated for each point in the mesh obtained. It is estimated from the ordered neighbours of the point. This stage can also be performed in a pipelined fashion {\it i.e.} once the neighbours $v$ is generated for a point $p$ its corresponding surface normal can be computed. There is no need to wait for the completion of the spin. A point forms a 3D vector when joined with its neighbour {\it i.e.} all the links in the mesh are actually vectors. For surface normal computation the direction of vector is towards its neighbours from the point in question. A normal can be estimated for a point $p$ if its corresponding $v$ has $\vert v \vert \geq 2 $. Thus, surface normal cannot be estimated for points with single link, though in reality that is a very rare scenario. Let the map $N=\{<p,n> \mid p \in P, n=\{\hat{i_p},\hat{j_p},\hat{k_p} \} \}$ stores the surface normal of all valid points. $\hat{i_p},\hat{j_p},\hat{k_p}$ are the normal components of point $p$. For valid points, the process of neighbour formation is described in Algorithm~\ref{algo:surfaceNormal}.

\begin{algorithm}[ht]
\label{algo:surfaceNormal}
\caption{Surface normal generation}
\SetAlgoLined
Let $N_{list} = <<\hat{i},\hat{j},\hat{k}>,w>$ be the temporary list of vectors\\
\For{all $p$ in cloud}{
Let $\vert v \vert = k_{n} $where $v$ is the neighbour vector of $p$\\
    \For{$i$ varied from $1$ to $k_{n}-1$ }{
     $\vec{A} = \overrightarrow{p,v_{i}}$
     $\vec{B} = \overrightarrow{p,v_{i+1}}$
     $\vec{C} = \vec{A} \times \vec{B}$\\
     $w=\frac{1}{max(\vert \vec{A} \vert,\vert \vec{B} \vert)}$\\
     $N_{list} \leftarrow <<\vec{C}_i,\vec{C}_j,\vec{C}_k>,w>$
      }
$\vec{A} = \overrightarrow{p,v_{k_n}}$
$\vec{B} = \overrightarrow{p,v_{1}}$
$\vec{C} = \vec{A} \times \vec{B}$\\
$w=\frac{1}{max(\vert \vec{A} \vert,\vert \vec{B} \vert)}$\\
$N_{list} \leftarrow <<\vec{C}_i,\vec{C}_j,\vec{C}_k>,w>$\\
$sum_i=0,sum_j=0,sum_k=0,sum_w=0$\\
\For{$t$ varied from $1$ to $k_{n}$}{
     $sum_i \leftarrow sum_i + (N_{list}(i_t)N_{list}(w_t))$\\
     $sum_j \leftarrow sum_j + (N_{list}(j_t)N_{list}(w_t))$\\
     $sum_k \leftarrow sum_k + (N_{list}(k_t)N_{list}(w_t))$\\
     $sum_w \leftarrow sum_w + N_{list}(w_t)$
    }
    $temp_n \leftarrow \{ \frac{sum_i}{sum_w},\frac{sum_j}{sum_w},\frac{sum_k}{sum_w} \} $\\
    $N \leftarrow <p,temp_n>$
}
\KwResult{N }
\end{algorithm}

To compute the surface normal at a point, it is connected to its neighbouring points in an anti-clockwise fashion to form a set of vectors. In the same order, the neighbouring vectors are cross multiplied to generate a set of candidate normals. Surface normal of a point is the weighted average of such candidates.
During the mesh formation the neighbours are stored in a sorted order as described in Section~\ref{subsec:mesh}. For example, let $\vec{A}$ and $\vec{B}$ are two consecutive vectors obtained by connecting the point with two neighbouring points. $\vec{C}$ formed as $\vec{A} \times \vec{B}$ is the corresponding candidate normal. The weight for  $\vec{C}$ is the inverse of the maximum of $\vert \vec{A} \vert$ and $\vert \vec{B} \vert$. Thus, in finding the weight of $\vec{C}$, neighbour at larger distance plays the major role and the weight is inversely proportional to the distance. Finally, candidates arising out of nearby neighbours will have more contribution towards the surface normal for the point. 
There are points in surface edges for which surface normal estimation is flawed due to interference of edge points from neighbouring surface, but it is almost impossible to distinguish between such surfaces at this stage of the algorithm.

\subsection{Segmentation by Surface Homogeneity}
\begin{figure}[ht]
\centering
\includegraphics[width=0.8\textwidth]{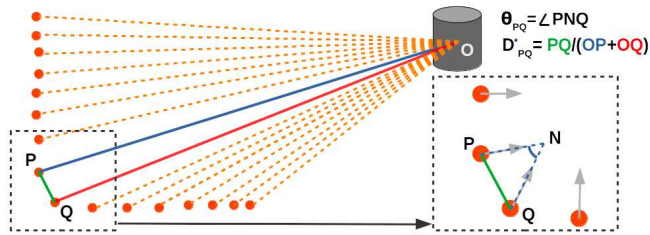}
\caption{Visual definition of $\theta_{pq}$ and $D^{*}_{pq}$. P and Q are shown at the juncture of two surfaces for a single vertical shot. }
\label{fig:seg_prop_logic}
\end{figure}

Once the surface normal of all the points are computed, surface segment proposals are generated using those normals. A label map $L=\{<p,l> \mid p \in P, l=0\}$ is used to assign a segment label $l$ to each point $p$. For any point $p$, initially $l=0$ denoting that the point is unlabelled. Whether two points $p$ and $q$ will get the same label or not depends on the angle $\theta_{pq}$ formed between their corresponding surface normal vectors and a normalized distance $D^{*}_{pq}$ between the points. Thus, surface homogeneity is determined by $\theta_{pq}$ and $D^{*}_{pq}$. Figure~\ref{fig:seg_prop_logic} visually elaborates the parameters for a single vertical shot. However, the relation will exist for all the links in the mesh. Segment labels are gradually increased and assigned to each segments. Considering the mesh as a graph, each segment label propagates following a depth first search approach. The algorithm for segment generation is elaborated in Algorithm~\ref{algo:surfaceSeg}. 
The normal map $N=\{<p,n> \mid p \in P, n=\{\hat{i_p},\hat{j_p},\hat{k_p} \} \}$ and the mesh $M=\{ <p,v> \mid p \in P, v = \{ q_n \mid q_n \in P, n\leq6,\quad and\quad q_n\quad is\quad a\quad neighbour\quad of\quad p \} \}$ are computed in the earlier stages. Subsequently Algorithm~\ref{algo:surfaceSeg} uses $N$ and $M$ to label the whole sub-sampled point cloud in an inductive fashion. 

\begin{algorithm}[ht]
\label{algo:surfaceSeg}
\caption{Surface segmentation based on the distribution of normal}
\SetAlgoLined
$L=\{<p,l> \mid p \in P, l=0\}$\;
$label\leftarrow 1, stack \: S\leftarrow\{\}$\;
\For{each $p \in M$}{
    \eIf{$l=0$ for $p$ in $L$}{
        $L \leftarrow <p,l\leftarrow label>$\;
        S.push(p)\;
        \While{$S \neq \{\}$}{
            $p\leftarrow$ S.pop()\;
            $\overrightarrow{p_{norm}} = <\hat{i_p},\hat{j_p},\hat{k_p}>$ for $p$ from $N$\;  
            \For{each $q$ in $v$ corresponding to $p \in M$}{
                \If{$l=0$ for $q$ in $L$}{
                    $\overrightarrow{q_{norm}} = < \hat{i_q},\hat{j_q},\hat{k_q}>$ for $q$ from $N$\;
                    $\theta_{pq}=cos^{-1}\frac{\overrightarrow{p_{norm}}\cdot \overrightarrow{q_{norm}}}{\vert \overrightarrow{p_{norm}} \vert \vert \overrightarrow{q_{norm}} \vert} $\;
                    $D^{*}_{pq}=\frac{\vert \overrightarrow{pq} \vert}{p_r+q_r}$\;
                    where $p_r$ and $q_r$ are the $r$ values of $p$ and $q$ respectively
                    
                    \If{$\theta_{pq}<\theta_{thres}$ and $D^{*}_{pq}<D^{*}_{thres}$}{
                        $L \leftarrow <q,l\leftarrow label>$\;
                        S.push(q)\;
                    }
                }
            }
        }
        $label \leftarrow label + 1$\;
    }
    {search next $p$ in $M$}
}
 
\KwResult{L }
\end{algorithm}

The algorithm starts at first point of the entire point cloud and proceeds in a column major order of the cloud expressed as a 2D array. The columns correspond to vertical shots of the Lidar during the sweep. For an interval factor of $k_{interval}$ the resultant subsampled array is of size $32 \times (1800/k_{interval})$ for a Lidar with $32$ laser units in the vertical array. Thus for interval of $5$ the resultant array is of size $32 \times 360$. All entries in the array will not be valid as many points are out of range and thus only points with $r$ within Lidar range are processed by the algorithm. The tree traversal may follow any direction to propagate. Once it can no more be spanned maintaining the surface homogeneity, the segment label is increased and the next unlabelled valid point becomes the seed. Thus a linear read is followed every time a new segment is generated. The number of segments will increase with stringent thresholds ({\it i.e.} small values for $\theta_{thres}$ and $D^{*}_{thres}$) and more complex scenes.  

The values $\theta_{thres}$ and $D^{*}_{thres}$ are chosen empirically to $0.2618$ (15 degrees in radian) and $0.05$ respectively. Due to sub-sampling, all points in $P$ will not be a part of any segment {\it i.e.} they will remain unlabelled. To generate a densely segmented map the labels of unlabelled points are estimated by their nearest labelled point in the same horizontal sweep. The horizontal sweep is chosen due to its high density of points. Due to spatial proximity the likelihood of getting the same label is much high along the horizontal sweep than the vertical array.

\subsection{Surface feature extraction and classification}
\label{subsec:feature}
After formation of surface segments they are to be classified to assign a semantic label.
A feature vector $f_l$ is formed for all unique labels $l$ in $L$. For a cloud the list of such vectors is stored in the map $F=\{<f_l,s_l> \forall \quad unique \quad l \in L \}$ where $s_l$ is the semantic class corresponding to all points with label $l$. For classification number of classifiers have been tried as discussed in Section~\ref{subsec:classifier}.
During training $s_l$ is supplied to the classifier and during testing the classifier reports the $s_l$ for an $f_l$. For all the segments the feature vectors $f_l$ are formed using Algorithm~\ref{algo:segSem}.

\begin{algorithm}[ht]
\label{algo:segSem}
\caption{Feature vector formation}
\SetAlgoLined
$F=\{<f_l,s_l> \forall \quad unique \quad l \in L \}$\;
$b \leftarrow $ bin size\;
Let $B(x)=\lfloor \frac{b(x+1)}{2} \rfloor \quad \forall \quad x<1$\;
$\quad \quad \quad \quad \: =b-1\quad for\quad x=1$\;

\For{all unique $l\in L$ }{
    $counter \leftarrow 0$\;
    $h_i[b] \leftarrow \{ 0 \}$,$h_j[b] \leftarrow \{ 0 \}$,$h_k[b] \leftarrow \{ 0 \}$\;
    \For{all $p\in N$ where $p\in \{ <p,l> \} $ }{
        $counter \leftarrow counter+1$\;
        $\overrightarrow{norm} \leftarrow n, <p,n> \in N$\;
        $h_i[B(\overrightarrow{norm}_i)]\leftarrow h_i[B(\overrightarrow{norm}_i)]+1$\;
        $h_j[B(\overrightarrow{norm}_j)]\leftarrow h_i[B(\overrightarrow{norm}_j)]+1$\;
        $h_k[B(\overrightarrow{norm}_k)]\leftarrow h_i[B(\overrightarrow{norm}_k)]+1$\;
    }
    normalize $h_i,h_j,h_k$\;
    $s_l\leftarrow argmax_{l}\{ \vert l \vert, l \in <p,l>, p\in N \}$ (for training only)\;
    $density\leftarrow \frac{counter}{\vert N \vert}$\;
    $f_l\leftarrow density\cdot h_i\cdot h_j\cdot h_k $ ($\cdot$ is concatenation)\;
    $F \leftarrow <f_l,s_l>$ (for training)\;
    $F \leftarrow f_l$ (for testing)\;
}

\KwResult{F }
\end{algorithm}

The concatenated histogram of surface normals of a segment along with the surface density form the feature to represent the semantic. Each of the surface normal histograms $h_i$, $h_j$ and $h_k$ is a distribution of the $\hat{i}$,$\hat{j}$ and $\hat{k}$ components respectively of all surface normals of a segment. The histograms are $b$-dimensional. For the present work, we have empirically set the value of $b$ as $16$ . Let $x$ stands for a normal component . Then, the function $B(x)$ as used in the algorithm determines the bin in the corresponding histogram $h_x$. The present work is concerned with semantic differentiation between different types of surface namely ``plane",``ground plane", ``cylinder", ``sphere" and ``cone". If an environment can be defined as a composition of such basic generator surfaces then with supervised combination of surfaces, complex models can be estimated. Also surface like ``ground plane" has immediate use in robot navigation. It can be observed from Algorithm~\ref{algo:surfaceSeg} that the histograms are normalized individually. 
This is done in order to prevent bias towards a particular component as the surface is composed of normals with variances differing for the components. The density factor also helps to bias the classifier towards correcting the labels of bigger segment proposals. During training the label of a feature vector is chosen as the one corresponding to majority of points in it. Due to local contextual surface propagation logic it may happen that different surfaces with a smooth transition may come under the same segment, the majority voting mitigates the effect of bad labelling in such cases.

The proposed methodology uses a statistical approach to prepare surface segment proposals and subsequently use them for semantic label prediction using classical classifier {\it i.e.} feature is not learnt rather engineered. This provides an insight into the structural properties of point cloud from spinning Lidar. Though deep classifiers are becoming popular, point cloud segmentation using them are computationally very expensive and requires sophisticated hardware. The proposed approach on the other hand can work on low configuration hardware providing decent accuracy, without sacrificing much on speed.

\section{Experimental Results and Comparison}
\label{sec:results_comp}
\begin{figure}[ht]
\centering
\includegraphics[width=\textwidth]{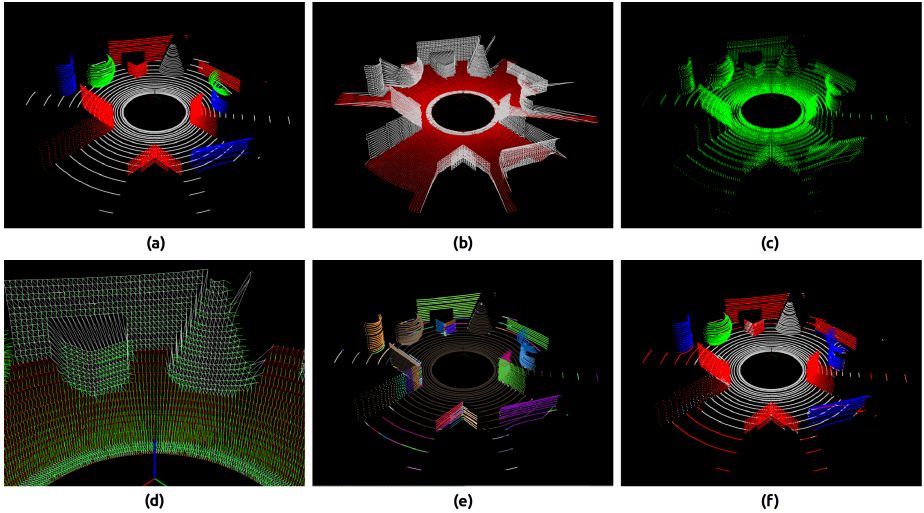}
\caption{Visual output at different stages (zoom in for better view): (a) cloud ground truth (b) resultant mesh (red portion is z filtered for better visualization) (c) surface normal on subsampled cloud (d) zoomed view of mesh and normal (e) segment proposals (f) semantic segmentation}
\label{fig:phase_shots}
\end{figure}

The proposed methodology is an updated and extended version of our previous work~\cite{mukherjee2019fast}. The system is realized using C++ language with OpenCV libraries for visualization purpose. The hardware configuration of the system used has 4GB DDR3 RAM and first generation Intel i5 processor. A synthetic dataset is prepared in order to test the methodology but the code can also run with live Lidar data stream from Velodyne Lidars. The visual output at different stages of the methodology is shown in Fig.~\ref{fig:phase_shots}. The surface segmentation results are compared  with the standard region growing algorithm used in point cloud library~\cite{Rusu2011} and a region growing algorithm combined with merging for organized point cloud data~\cite{Zhan2009}. The semantic segmentation results are compared with pointnet~\cite{Qi2017} and pointnet++~\cite{Qi2017pnpp}.

\begin{figure}[ht]
\centering
\includegraphics[width=\textwidth]{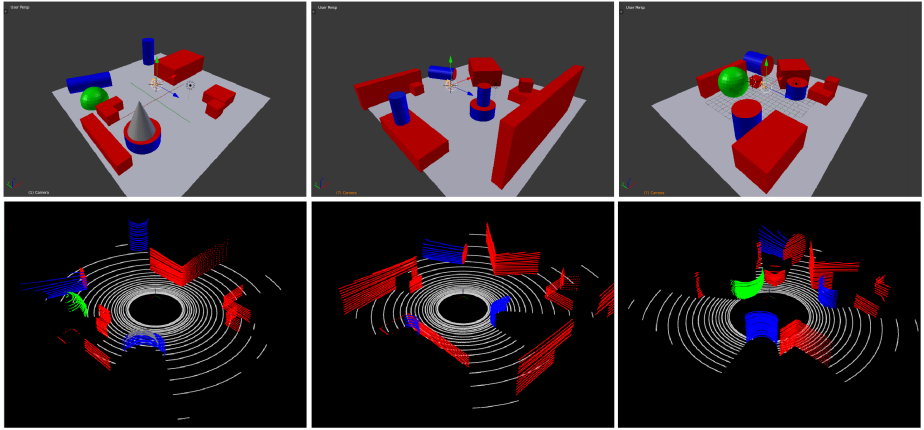}
\caption{Sample point cloud for different scenes: top row shows the scenes created using Blensor, bottom row shows the corresponding clouds.} 
\label{fig:dataset_samples}
\end{figure}

\subsection{Synthetic dataset}
A synthetic dataset is created using the ``Blensor" tool~\cite{Gschwandtner2011}. Environmental model files were created that contains regular shaped objects in different orders of scale, orientation, density and occlusion. Four kinds of surfaces are placed on the scene, namely ``plane", ``cylinder", ``sphere" and ``cone". The ground plane is labelled as different surface namely ``ground plane". A Velodyne 32E Lidar was simulated with $0.2$ degree horizontal resolution, thus producing a maximum possible point cloud of size $32 \times 1800$ for a scene. A Gaussian noise model with zero mean and variance of $0.01$ is incorporated in the sensor. It must be admitted that in reality spinning Lidars are more accurate. However, here the extra noise is incorporated to test the robustness of the methodology. There are $32$ unique environments (scenes). This is the {\it primary dataset} on which we have tested our surface segmentation process ({\it i.e.} prior to semantic label assignment). We refer to this as non semantic segmentation. At this stage only the surfaces are extracted. Figure~\ref{fig:dataset_samples} shows some sample scenes along with their point clouds. The different surfaces of ``plane", ``ground plane", ``cylinder", ``sphere" and ``cone" are colored as red, off white, blue, green and grey respectively. The color code holds for all other ground truths and semantic output.

In order to put the semantic label, we rely on classifiers. Classifiers are to be trained with sufficient data. The data should correspond to a good mix of different types of surfaces. In our primary dataset $8$ scenes are having only one object and mostly contain ground plane. It may bias the training. Hence we have considered the remaining $24$ scenes that are comparatively complex with multiple objects. Thus, corresponding cloud will have good proportion of different types of surfaces. We need to augment the dataset also. Hence, for semantic labelling, $289$ unique clouds are produced from $24$ unique scenes, by shifting the Lidar horizontally. Then a random selection is done in order to divide it in a training set consisting of $173$ clouds, validation set of $29$ clouds and testing set of $87$ clouds. 
The training set is further augmented to $692$ clouds by mirroring each of the $173$ point cloud along the $x$ and $y$ axes. For semantic labelling we consider this dataset and will refer this one as {\it semantic dataset}.

\subsection{Comparison of performance: non semantic segmentation}
For non semantic segmentation, the performance of the proposed methodology is compared with the others using the precision, recall and f1 score metrics. The evaluation has been performed on the {\it primary dataset} as discussed earlier. Every individual surface in the scene is annotated by a marker/number (not semantically labelled). Hence it is difficult to compare the output of a methodology with the groundtruth by matching the number/marker. Hence, an edge based matching is performed. The overlap of dilated ground truth edges with the edges reported by the methodologies have been compared. 

To reduce the number of points in the cloud, sampling is done as discussed in Section~\ref{sec:method}. In our experiment, $k_{interval}$ has been varied from $5$ to $15$ in steps of $5$ (note, an interval of $5$ corresponds to $1$ degree of horizontal sweep of the Lidar). The thresholds for angular difference and normalized distance are kept at $\theta_{thres}=0.2618$ ($15$ degree in radian) and $D^{*}_{thres}=0.05$ respectively. Our methodology is compared with~\cite{Rusu2011} and~\cite{Zhan2009}. The tuning parameters of these methods are kept at the default settings as suggested by the original authors. Table~\ref{tab:nonSem_time} shows the comparative latency and Table~\ref{tab:nonSem_acc} shows the comparative accuracy of the different methodologies. It can be said that the proposed methodology for non semantic segmentation fares significantly well in comparison to others both, in terms of speed and accuracy. It is observed that for sampling interval of $5$ the proposed methodology gives best accuracy without much compromise in speed and thus it is kept at this value to study the performance further. 

\begin{table}[ht]
\centering
\begin{tabular}{|c|c|c|c|c|}
\hline
Methodology & \begin{tabular}[c]{@{}c@{}}Sampling \\ Interval\end{tabular} & \begin{tabular}[c]{@{}c@{}}Average \\ Time (in ms)\end{tabular} & \begin{tabular}[c]{@{}c@{}}Max \\ Time (in ms)\end{tabular} & \begin{tabular}[c]{@{}c@{}}Min \\ Time (in ms)\end{tabular} \\ \hline
\multirow{3}{*}{\begin{tabular}[c]{@{}c@{}}Proposed \\ Methodology\end{tabular}} & 5 & 56.67 & 68 & 43 \\ \cline{2-5} 
 & 10 & 36.50 & 52 & 31 \\ \cline{2-5} 
 & 15 & 26.85 & 36 & 21 \\ \hline
\multicolumn{2}{|c|}{Region Growing~\cite{Rusu2011}} & 290.50 & 1588 & 141 \\ \hline
\multicolumn{2}{|c|}{\begin{tabular}[c]{@{}c@{}}Region Growing with\\ Merging~\cite{Zhan2009}\end{tabular}} & 377.68 & 1732 & 132 \\ \hline
\end{tabular}
\caption{Comparison of latency of different methods}
\label{tab:nonSem_time}
\end{table}

\begin{table}[ht]
\centering
\begin{tabular}{|c|c|c|c|c|}
\hline
Methodology & \begin{tabular}[c]{@{}c@{}}Sampling \\ Interval\end{tabular} & \begin{tabular}[c]{@{}c@{}}Average \\ F1 Score\end{tabular} & \begin{tabular}[c]{@{}c@{}}Average\\ Precision\end{tabular} & \begin{tabular}[c]{@{}c@{}}Average\\ Recall\end{tabular} \\ \hline
\multirow{3}{*}{\begin{tabular}[c]{@{}c@{}}Proposed \\ Methodology\end{tabular}} & 5 & 0.78 & 0.80 & 0.78 \\ \cline{2-5} 
 & 10 & 0.75 & 0.75 & 0.75 \\ \cline{2-5} 
 & 15 & 0.71 & 0.74 & 0.69 \\ \hline
\multicolumn{2}{|c|}{Region Growing~\cite{Rusu2011}} & 0.35 & 0.48 & 0.28 \\ \hline
\multicolumn{2}{|c|}{\begin{tabular}[c]{@{}c@{}}Region Growing with\\ Merging~\cite{Zhan2009}\end{tabular}} & 0.36 & 0.39 & 0.34 \\ \hline
\end{tabular}
\caption{Comparison of accuracy of different methods}
\label{tab:nonSem_acc}
\end{table}

\subsection{Choice of classifier}
\label{subsec:classifier}
For semantic classification of the extracted surfaces, traditional classifiers have been tried and their performances are evaluated. Section~\ref{subsec:feature} discussed about the features used for classification. We have worked with {\it semantic dataset} for semantic labelling. The classifiers have been trained with the augmented training set. The validation set has been used to evaluate the performance of different classifiers. Finally, a classifier is chosen based on the comparison metrics obtained on the validation set. Classifiers evaluated are multimodal Support Vector Machine with RBF kernel~\cite{lee2004svm}, K Nearest Neighbour classifier~\cite{altman1992knn}, Decision Tree~\cite{quinlan1986dt}, Random Decision Forest~\cite{ho1995} and Extremely Randomized Tree~\cite{geurts2006}. Table~\ref{tab:comp_class} shows the comparative F1 score, precision and recall.
As all the classifiers perform in the similar manner with respect to latency, the accuracy became the important parameter in making the judgement. Even in terms of the accuracy parameters also all the classifiers provide reasonably good outcome. It indicates the strength of the proposed feature vector. Finally, based on the accuracy, we consider either of the top two classifiers {\it i.e.} random decision forest or extremely randomized tree can be selected for semantic labelling. 

\begin{table}[ht]
\centering
\begin{tabular}{|c|c|c|c|}
\hline
Classifier & \begin{tabular}[c]{@{}c@{}}Average\\ F1 score\end{tabular} & \begin{tabular}[c]{@{}c@{}}Average \\ precision\end{tabular} & \begin{tabular}[c]{@{}c@{}}Average\\ recall\end{tabular} \\ \hline
\begin{tabular}[c]{@{}c@{}}Multiclass SVM\\ with RBF kernel~\cite{lee2004svm}\end{tabular} & 0.68 & 0.66 & 0.71 \\ \hline
\begin{tabular}[c]{@{}c@{}}K Nearest \\ Neighbour~\cite{altman1992knn}\end{tabular} & 0.73 & 0.72 & 0.73 \\ \hline
Decision tree~\cite{quinlan1986dt} & 0.70 & 0.70 & 0.70 \\ \hline
\begin{tabular}[c]{@{}c@{}}Random Descision\\ Forest (RDF)~\cite{ho1995}\end{tabular} & 0.74 & 0.75 & 0.74 \\ \hline
\begin{tabular}[c]{@{}c@{}}Extremely Randomized\\ Trees (ERT)~\cite{geurts2006}\end{tabular} & 0.77 & 0.77 & 0.76 \\ \hline
\end{tabular}
\caption{Comparative accuracy for semantic labelling by different classifiers on the validation set}
\label{tab:comp_class}
\end{table}

\subsection{Comparison of performance: semantic segmentation}

\begin{figure}[ht]
\centering
\includegraphics[width=\textwidth]{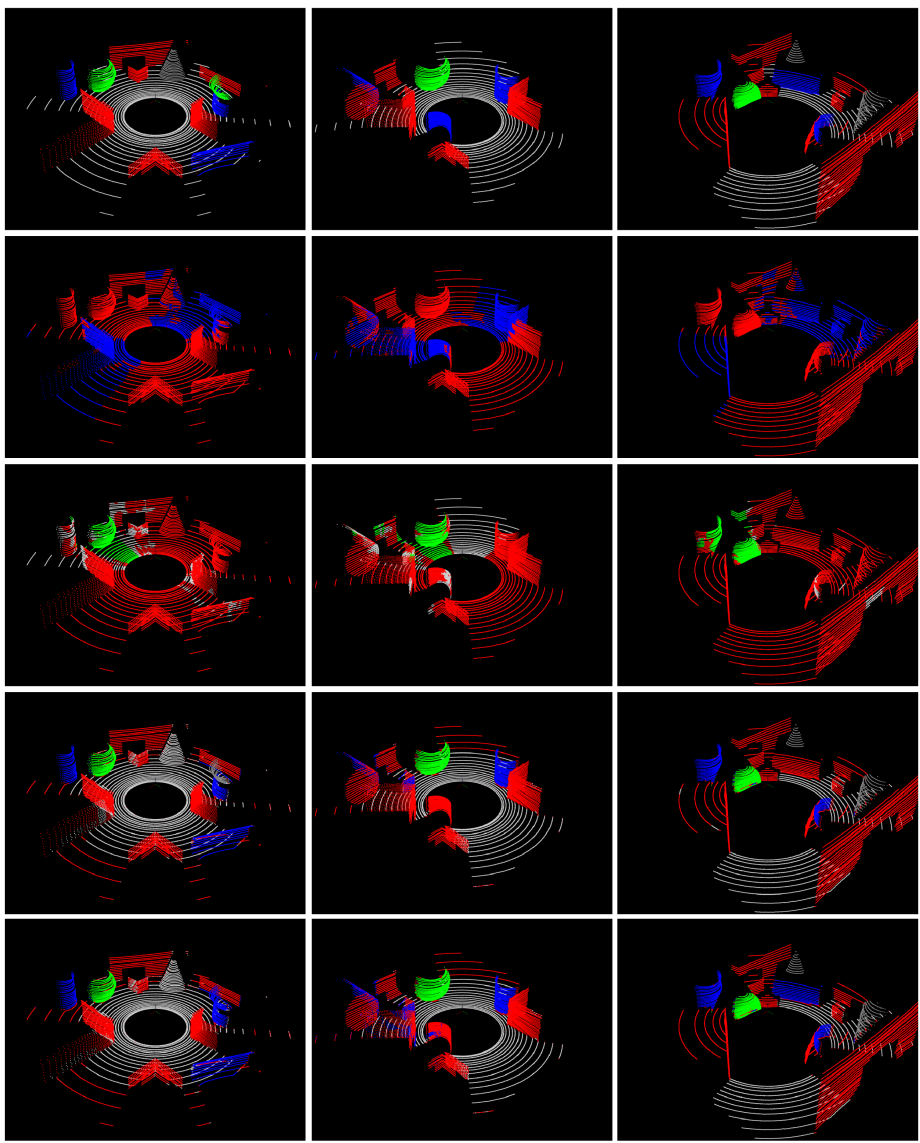}
\caption{Output of different methods. From top to bottom rows correspond to ground truth of three scenes, output of pointnet~\cite{Qi2017}, output of pointnet++~\cite{Qi2017pnpp}, output of proposed methodology with RDF classifier, output of proposed methodology with ERT classifier respectively}
\label{fig:comparitive_shots}
\end{figure}

For semantic segmentation the performance of the proposed methodology is compared with others using the mean intersection over union (MIoU) metric. As both ground truth and experimental output points belongs to a definite semantic class, such a comparison is possible. Classwise and overall comparison in terms of MIOU are made considering all the point clouds in the test set. Average precision, recall and F1 scores are also provided. The proposed methodology is compared with the pointnet~\cite{Qi2017} and pointnet++~\cite{Qi2017pnpp}. These deep learning based methods are executed in a Linux machine with Intel Xeon 2.3GHz processor, Tesla K80 12GB GPU and 128GB DDR4 RAM. The pointnet and pointnet++ networks are designed to process small dense point clouds. On the other hand our dataset consists of large sparse clouds. Hence, contiguous points of size $1038$ are fed to the network at one shot during both, training and testing. This is due to the constraint posed by GPU memory. The methods works with local contextual information and thus breaking the clouds into smaller chunks does not interfere with its working principle. Figure~\ref{fig:comparitive_shots} shows the output of different methods corresponding to few sample scenes. It can be observed that output of proposed methodology (for both the classifiers) is better than the others for all the classes. Although classes like ``plane" and ``ground plane" are well detected by the proposed methodology, detection of ``cone" suffers.  

Table~\ref{tab:semSeg_time} shows the relative latency of the methods. Table~\ref{tab:semSeg_acc_f1} shows the relative accuracy of the methods in terms of average F1 score, precision and recall over all points in all clouds of the test set. Table~\ref{tab:semSeg_acc_miou} shows the class-wise and overall accuracy of the methods in terms of MIoU. For calculating the overall MIoU all points from all clouds are used rather than average of MIoU for individual test clouds.

\begin{table}[ht]
\centering
\begin{tabular}{|c|c|c|c|}
\hline
Methodology & \begin{tabular}[c]{@{}c@{}}Average\\ Time (in ms)\end{tabular} & \begin{tabular}[c]{@{}c@{}}Maximum\\ Time (in ms)\end{tabular} & \begin{tabular}[c]{@{}c@{}}Minimum\\ Time (in ms)\end{tabular} \\ \hline
Pointnet~\cite{Qi2017} & 176.36 & 359 & 76 \\ \hline
Pointnet++~\cite{Qi2017pnpp} & 2462.65 & 2541 & 2056 \\ \hline
\begin{tabular}[c]{@{}c@{}}Proposed Methodology\\ with RDF classifier\end{tabular} & 109 & 124 & 96 \\ \hline
\begin{tabular}[c]{@{}c@{}}Proposed Methodology\\ with ERT classifier\end{tabular} & 98 & 115 & 82 \\ \hline
\end{tabular}
\caption{Comparative latency of different methodologies for semantic segmentation}
\label{tab:semSeg_time}
\end{table}

\begin{table}[ht]
\centering
\begin{tabular}{|c|c|c|c|}
\hline
Methodology & \begin{tabular}[c]{@{}c@{}}Average\\ F1 score\end{tabular} & \begin{tabular}[c]{@{}c@{}}Average\\ Precision\end{tabular} & \begin{tabular}[c]{@{}c@{}}Average\\ Recall\end{tabular} \\ \hline
Pointnet~\cite{Qi2017} & 0.28 & 0.28 & 0.28 \\ \hline
Pointnet++~\cite{Qi2017pnpp} & 0.41 & 0.41 & 0.42 \\ \hline
\begin{tabular}[c]{@{}c@{}}Proposed Methodology\\ with RDF classifier\end{tabular} & 0.76 & 0.77 & 0.76 \\ \hline
\begin{tabular}[c]{@{}c@{}}Proposed Methodology\\ with ERT classifier\end{tabular} & 0.77 & 0.77 & 0.78 \\ \hline
\end{tabular}
\caption{Overall accuracy (in terms of recall, precision and F-measure) of semantic segmentation by different methodologies}
\label{tab:semSeg_acc_f1}
\end{table}

\begin{table}[ht]
\centering
\begin{tabular}{|c|c|c|c|l|l|l|}
\hline
Methodology & Plane & \begin{tabular}[c]{@{}c@{}}Ground \\ Plane\end{tabular} & Sphere & Cylinder & Cone & Overall \\ \hline
Pointnet~\cite{Qi2017} & 0.32 & 0.00 & 0.00 & 0.10 & 0.00 & 0.17 \\ \hline
Pointnet++~\cite{Qi2017pnpp} & 0.40 & 0.38 & 0.36 & 0.16 & 0.00 & 0.39 \\ \hline
\begin{tabular}[c]{@{}c@{}}Proposed Methodology \\ with RDF classifier\end{tabular} & 0.66 & 0.64 & 0.46 & 0.29 & 0.18 & 0.62 \\ \hline
\begin{tabular}[c]{@{}c@{}}Proposed Methodology\\ with ERT classifier\end{tabular} & 0.68 & 0.64 & 0.45 & 0.31 & 0.19 & 0.63 \\ \hline
\end{tabular}
\caption{Class wise and overall accuracy (in terms of MIoU) of semantic segmentation for different methodologies}
\label{tab:semSeg_acc_miou}
\end{table}

From analysis of comparative results it can be said that the proposed methodology can deliver acceptable accuracy at real time speed. In general, spinning Lidars are operated at a maximum speed of $10$ rotations per seconds and the average time of execution of our methodology is $115$ to $124$ milliseconds ($8$ to $8.7$ FPS) depending on the choice of classifier. As a portion of the methodology is computed along with the Lidar spin, with a more optimized version the system can run in real time without any frame loss. Thus the proposed methodology can run on non-GPU low configuration system, in real time, delivering an MIoU accuracy of over $60\%$.

\section{Conclusion}
\label{sec:conc}
The present work deals with the problem of semantic surface segmentation from Lidar point cloud data. The proposed methodology has a novel fast meshing process that generates surface mesh from the Lidar scan in an online fashion, facilitating fast computation of surface normals. Subsequently a statistical method generates segment proposals. The proposals are described with a novel feature vector based on the distribution of surface normals. Semantic labelling is done by feeding the feature vector as input to classifier.
The performance of the proposed methodology is compared with some popular cloud segmentation methods. It is observed that the proposed methodology is significantly faster and provides higher classification accuracy. 
It can be concluded that the proposed methodology can deliver acceptable accuracy for robotic applications in real time and paves the way for further utilization of semantic surfaces towards generation of models and scene reconstruction.


%
%

\bibliography{mybibliography}   

%
%

\end{document}